# Incremental Transformer for Surrogate-Based Inverse Design of Geopolymer Mixtures

Giansalvo Cirrincione[1], Filippo Grassia[1*]
[1]Laboratory of Innovative Technologies (LTI, EA 3899)
University of Picardie Jules Verne, Amiens, 80025, Hauts-de-France, France

**Abstract**

Small-data inverse design is a recurrent challenge in engineering informatics, particularly when observations are heterogeneous, mixed-type, and constrained by physical relations among design variables. A topology-aware surrogate framework guided by an Incremental Transformer (INCRT) is proposed for physics-constrained inverse design, applied to geopolymer mixture design. The method integrates intrinsic-dimensionality analysis, mixed-variable design-space representation, tabular surrogate prediction, INCRT-based manifold rationalisation, and constrained inverse optimisation.

The case study uses a public benchmark of fly-ash and ground-granulated-blast-furnace-slag geopolymer concrete mixtures with compressive-strength and carbon-emission targets. The high-dimensional design space proves strongly redundant, organising around a smaller number of effective mixture regimes. Forward prediction shows that compressive strength requires nonlinear tabular surrogates, whereas carbon emission is largely determined by mixture composition and recovered accurately by regularised linear models. INCRT therefore serves not as a replacement for strong tabular predictors, but as a topology-aware rationalisation layer supplying prototype regimes and a manifold-support score for inverse design.

Three inverse-design strategies are compared: unconstrained surrogate optimisation, physics-constrained optimisation, and topology-aware physics-constrained optimisation. Unconstrained optimisation can match target strength but may yield physically invalid or off-manifold candidates; physics-only constraints do not always guarantee data support. The topology-aware strategy produces candidate mixtures balancing target compliance, carbon reduction, physical admissibility, and proximity to the learned feasible manifold.

The framework is not intended to replace experimental validation, but to provide a decision-support methodology for screening credible candidate mixtures from small, mixed, physically constrained engineering datasets.

## 1 Introduction

Many engineering design problems are increasingly supported by data-driven models, yet the datasets available in practical design contexts often remain small, heterogeneous, and physically constrained. This situation is common in material mixture design, where each experiment may require laboratory preparation, curing, destructive testing, and environmental assessment. As a consequence, the available data rarely cover the design space uniformly. Observations usually lie on a restricted subset of physically meaningful recipes, shaped by chemistry, processing constraints, and prior engineering practice.

*Corresponding author: Filippo Grassia, filippo.grassia@u-picardie.fr

Geopolymer concrete (GPC) is a representative example. It is widely investigated as a lower-carbon alternative to ordinary Portland cement concrete because aluminosilicate precursors such as fly ash (FA) and ground granulated blast-furnace slag (GGBFS) can be activated by alkaline solutions to form hardened binders (Davidovits, 2015; Provis and van Deventer, 2009). The mechanical performance of GPC mixtures depends on nonlinear interactions among precursor composition, activator chemistry, water content, curing temperature, curing time, and aggregate proportions. At the same time, environmental indicators such as carbon dioxide emission are often computed from constituent quantities and emission factors, and therefore behave more like composition-driven quantities.

Most data-driven studies on GPC mixtures focus on forward prediction, namely estimating compressive strength or related properties from mixture descriptors Nguyen et al. (2025); Mohsin et al. (2025). Forward prediction is necessary, but it is not sufficient for engineering inverse design. In inverse design, a surrogate model is embedded inside an optimiser in order to generate new candidate mixtures satisfying target requirements. This changes the problem substantially. A surrogate may interpolate accurately within the sampled data distribution but behave unreliably when queried in regions that are sparsely represented or outside the observed design manifold. This is particularly problematic in small-data settings, where many numerical optima may be artefacts of extrapolation.

A candidate mixture should therefore not be considered valid merely because it optimises a surrogate objective. It must also satisfy physical constraints and remain sufficiently close to the feasible design manifold inferred from the data. This paper addresses this issue by proposing an INCRT-guided topology-aware surrogate inverse-design framework. INCRT (Incremental Transformer) is a self-organising architecture that determines its own structure during training, originally introduced by Cirrincione (unpublished results). In the present paper, INCRT is used in a reduced operational form: the prototype set produced by the architecture is exploited to organise mixtures into prototype-supported regimes and to provide a manifold-support score. The role of INCRT is not to replace strong tabular predictors in forward prediction. Instead, it acts as a rationalisation and out-of-distribution control layer. Strong surrogate models are used for compressive-strength and carbon-emission prediction, while the INCRT/topology layer constrains inverse design toward physically admissible and data-supported mixture regimes.

The resulting workflow is summarised in figure 1. It starts from a small mixed tabular dataset, constructs physically interpretable descriptors, analyses intrinsic dimensionality, trains accurate forward surrogates, extracts topology-aware prototypes, and finally compares unconstrained, physics-constrained, and topology-aware inverse design.

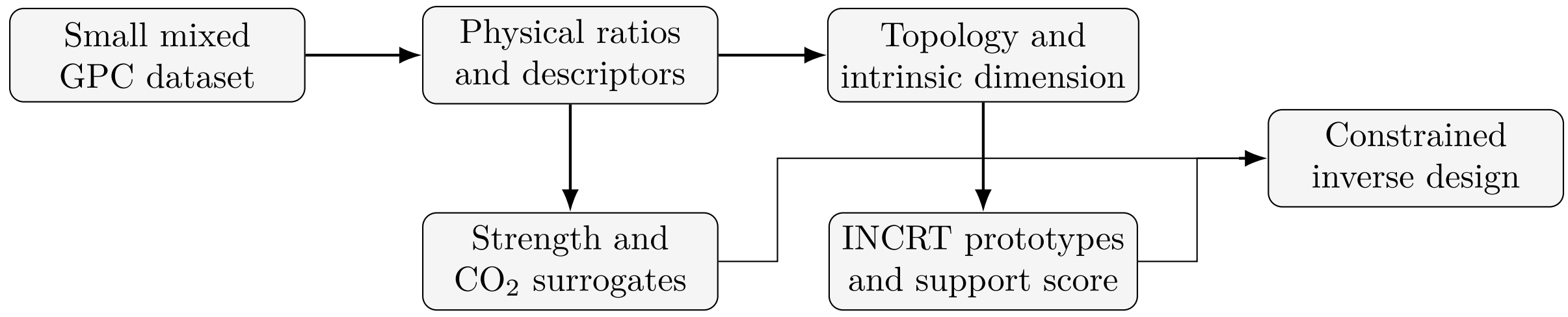

**Figure 1:** Workflow.

For readability, the method can be understood as a sequence of four filters applied to a possible recipe, as shown in table 1. The first filter is numerical prediction: the recipe must have an acceptable predicted strength and carbon emission. The second filter is physical admissibility: material quantities, ratios, and curing levels must remain plausible. The third filter is topological support: the candidate should not be isolated from the mixture regimes observed in the data. The fourth filter is engineering interpretation: a small strength tolerance may be acceptable if it produces a substantial carbon reduction and remains inside the data-supported region. This layered interpretation is used throughout the paper to avoid treating inverse design as a purely numerical optimisation problem.

**Table 1:** Design filters.

| Filter | Question answered | Failure mode avoided |
|---|---|---|
| Prediction | Does the surrogate predict the requested performance? | Candidate has poor predicted strength or carbon score. |
| Physical admissibility | Does the recipe respect basic mixture and process constraints? | Candidate is numerically attractive but chemically or procedurally implausible. |
| Manifold support | Is the candidate close to observed prototype regimes? | Candidate is an extrapolative surrogate artefact. |
| Engineering interpretation | Is the trade-off acceptable for screening? | Exact target matching is overvalued at the expense of credibility. |

The main contribution of this paper is the formulation of an INCRT-guided topology-aware surrogate framework for inverse design from small, mixed, physically constrained engineering datasets. Accurate surrogate models are used for compressive-strength and carbon-emission prediction, while INCRT provides a prototype-based rationalisation of the feasible design manifold and a manifold-support score for inverse optimisation. This makes it possible to show that candidate mixtures should be selected not only by predicted performance, but also by physical admissibility and data support.

The specific contributions are as follows. First, GPC mixture design is formulated as a small-data, mixed-variable, physics-constrained inverse-design problem. Second, the 19-dimensional tabular representation is shown to be highly redundant through principal component analysis (PCA), participation ratio, entropy effective rank, local intrinsic-dimensionality estimates, manifold embedding, and clustering. Third, a hybrid surrogate-INCRT architecture is introduced, where ExtraTrees and related tree ensembles support nonlinear strength prediction, regularised linear models support carbon-emission prediction, and INCRT prototypes support out-of-distribution control. Fourth, repeated stratified cross-validation, nested model selection, and leave-one-cluster-out validation are combined to distinguish interpolation from regime-shift generalisation. Fifth, three inverse-design strategies are compared: unconstrained surrogate optimisation, physics-constrained optimisation, and topology-aware physics-constrained optimisation.

The remainder of the paper is organised as follows. Related work is reviewed in section 2. The inverse-design problem and notation are introduced in section 3. The dataset and feature construction are described in section 4. The topology-aware method and validation protocol are presented in section 5. Results are reported in section 6. The implications, limitations, and conclusion are given in sections 7 and 8.

## 2 Related work

Data-driven inverse design has become an important topic in engineering informatics, materials informatics, and computational mixture design. In a common setting, a surrogate model is trained from available experimental or simulation data and then used inside an optimisation loop to identify candidate designs satisfying one or more target properties. This approach is attractive when direct experimentation or high-fidelity simulation is expensive. However, inverse design differs from standard supervised prediction. In supervised prediction, model quality is usually evaluated on held-out test data sampled from the same distribution as the training data. In inverse design, the optimiser may actively search regions that are poorly represented in the training set. High cross-validation accuracy does not necessarily imply reliable design performance.

A useful distinction is between interpolation, controlled extrapolation, and unsupported extrapolation. Interpolation occurs when a candidate lies among previously observed mixtures. Controlled extrapolation occurs when a candidate moves slightly beyond known recipes while preserving physical consistency and neighbourhood support. Unsupported extrapolation occurs when the optimiser reaches a region that is numerically allowed but not represented by the data. Standard cross-validation mainly evaluates interpolation. Inverse design can easily enter the third regime. The present framework is designed to make this distinction explicit.

Machine learning has been widely applied to the prediction of geopolymer properties, especially compressive strength. The importance of oxide composition, precursor chemistry, and curing variables has been emphasised in the geopolymer literature (Reddy et al., 2016; Davidovits, 2015; Provis and van Deventer, 2009). More recently, GPC and alkali-activated-material databases have been combined with machine-learning methods to support prediction and low-carbon mixture design (Pham and Nguyen, 2024; Völker et al., 2023). The dataset considered in this paper is the public Mendeley dataset of Pham (2023), which contains 274 data points, 19 input variables, compressive strength, and carbon-emission information for FA/GGBFS-based GPC mixtures. A related journal study by Pham and Nguyen (2024) uses the same general modelling context to analyse the effects of alkaline activators and other factors on GPC properties with gene-expression-programming-based models. This connection is useful because it confirms that the dataset is not only a numerical table, but also a chemically structured mixture-design problem.

Existing machine-learning studies often emphasise predictive accuracy under random train-test splits. Such splits are useful for interpolation, but they may overestimate reliability when similar mixture regimes appear in both training and test sets. Moreover, direct optimisation of surrogate predictions may generate recipes far from experimentally observed regimes. The present work therefore treats forward prediction as only one part of a broader inverse-design problem.

Low-carbon design of construction materials requires balancing mechanical performance and environmental impact. In concrete and GPC design, carbon emission is often estimated from constituent quantities and emission factors. This makes carbon emission a largely composition-driven target, whereas strength depends on nonlinear material-formation mechanisms. Multi-objective formulations are therefore natural: candidate mixtures may be required to reach a target compressive strength while reducing carbon emission, cost, curing energy, or other sustainability indicators. However, multi-objective optimisation increases the risk of surrogate extrapolation. Low-carbon optima may be obtained by moving toward extreme compositions that are not experimentally represented. This provides an additional motivation for topology-aware constraints.

Prototype-based and topology-aware learning provide useful tools in this setting. PCA is used to diagnose dominant variance directions (Jolliffe, 2002). Entropy effective rank and participation-ratio-type measures quantify effective dimensionality (Roy and Vetterli, 2007). Local intrinsic dimensionality can be estimated from nearest-neighbour distances (Levina and Bickel, 2005). Cluster validity can be assessed by silhouette analysis (Rousseeuw, 1987). These tools do not replace physical modelling, but they help determine whether the observed mixtures occupy a thin subset of the ambient feature space. When this is the case, inverse design should respect that structure.

The present work is positioned at the intersection of small-data engineering informatics, geopolymer mixture modelling, topology-aware learning, and constrained inverse design. Its distinctive feature is the explicit separation between prediction and design support. Accurate surrogate models are selected empirically for strength and carbon-emission prediction, while INCRT-based topology is used to constrain the inverse-design search to data-supported regions.

**Table 2:** Positioning of the study.

| Line of work | Main focus | Typical limitation | Position of this paper |
|---|---|---|---|
| Geopolymer ML | Forward prediction of material properties | Limited inverse-design control | Forward prediction is embedded in constrained design. |
| Low-carbon design | Strength-emission trade-off | Possible surrogate extrapolation | Physical and manifold-support constraints are added. |
| Surrogate optimisation | Numerical design search | Optimiser may exploit model artefacts | Off-manifold candidates are penalised. |
| Prototype learning | Local representatives and interpretation | Not always the best predictor | Prototypes are used for rationalisation and support. |

The positioning in table 2 also clarifies the reason for not framing the paper as a simple benchmark competition. A model that obtains the lowest root-mean-square error is valuable, but it does not by itself define a safe inverse-design procedure. Conversely, a prototype model that is not the best predictor can still be valuable if it identifies supported regions of the design space. This distinction is central for small engineering datasets, where the cost of an implausible recommendation is higher than the cost of a slightly conservative candidate.

# 3 Problem formulation and notation

All variables used in the mathematical formulation are introduced before the corresponding equations. Let $n$ denote the number of available mixtures and let $d$ denote the number of original input variables. In the present dataset, $n = 274$ and $d = 19$. Let $\mathcal{X} \subset \mathbb{R}^d$ be the admissible ambient design domain, and let $x_i \in \mathcal{X}$ be the descriptor vector of mixture $i$. The dataset is denoted by

$$\mathcal{D} = \{(x_i, y_i)\}_{i=1}^{n}, \tag{1}$$

where $y_i$ is the measured or computed response associated with $x_i$. The compressive-strength target is denoted by $y_i^{\mathrm{str}}$, and the carbon-emission target is denoted by $y_i^{\mathrm{CO_2}}$. If both targets are used, the response vector is

$$y_i = \left(y_i^{\mathrm{str}}, y_i^{\mathrm{CO_2}}\right). \tag{2}$$

The design vector may contain continuous quantities, discrete process variables, and indicator-like variables. These components are denoted respectively by $x_i^{(c)}$, $x_i^{(d)}$, and $x_i^{(b)}$. Therefore,

$$x_i = \left(x_i^{(c)}, x_i^{(d)}, x_i^{(b)}\right). \tag{3}$$

This decomposition is used only to define appropriate distances and projections; it does not imply that the original dataset must already be separated in this form.

Let $f_{\mathrm{str}}$ be the surrogate model used to predict compressive strength and let $f_{\mathrm{CO_2}}$ be the surrogate model used to predict carbon emission. For a prescribed strength target $y^\star$, the simplest inverse-design objective would minimise the squared target error,

$$\mathcal{J}_{\mathrm{str}}(x) = \left(f_{\mathrm{str}}(x) - y^\star\right)^2. \tag{4}$$

However, equation (4) ignores physical feasibility and data support. Let $\Phi_{\mathrm{phys}}(x)$ be a non-negative penalty measuring violation of physical or procedural constraints. Let $\lambda_{\mathrm{phys}} \geq 0$ be its weight. A physics-constrained objective is

$$\mathcal{J}_{\mathrm{phys}}(x) = \left(f_{\mathrm{str}}(x) - y^\star\right)^2 + \lambda_{\mathrm{phys}}\Phi_{\mathrm{phys}}(x). \tag{5}$$

For carbon-aware design, let $\alpha \geq 0$ and $\beta \geq 0$ be the weights of strength-target error and carbon emission. The corresponding objective is

$$\mathcal{J}_{\mathrm{multi}}(x) = \alpha\left(f_{\mathrm{str}}(x) - y^\star\right)^2 + \beta f_{\mathrm{CO_2}}(x) + \lambda_{\mathrm{phys}}\Phi_{\mathrm{phys}}(x). \tag{6}$$

The topology-aware formulation introduces a prototype set $\mathcal{P} = \{p_k\}_{k=1}^{K}$, where $p_k$ denotes the $k$-th representative mixture regime and $K$ is the number of prototypes. Let $d_{\text{mix}}(x, z)$ be a mixed-variable distance between two candidates $x$ and $z$. This distance combines continuous, discrete, and binary components:

$$d_{\text{mix}}(x, z) = \alpha_c \left\| x^{(c)} - z^{(c)} \right\|^2 + \alpha_d d_d \left( x^{(d)}, z^{(d)} \right) + \alpha_b d_b \left( x^{(b)}, z^{(b)} \right), \tag{7}$$

where $\alpha_c$, $\alpha_d$, and $\alpha_b$ are non-negative scaling coefficients; $d_d$ is a distance for discrete variables; and $d_b$ is a distance for binary indicators. The manifold-support score is defined as

$$\Phi_{\text{man}}(x) = \min_{p_k \in \mathcal{P}} d_{\text{mix}}(x, p_k)^2. \tag{8}$$

Small values of $\Phi_{\text{man}}(x)$ indicate that $x$ lies close to a prototype-supported region. Large values indicate that $x$ is far from all observed regimes. With a non-negative manifold weight $\lambda_{\text{man}}$, the final topology-aware objective is

$$\mathcal{J}_{\text{top}}(x) = \alpha \left( f_{\text{str}}(x) - y^{\star} \right)^2 + \beta f_{\text{CO}_2}(x) + \lambda_{\text{phys}} \Phi_{\text{phys}}(x) + \lambda_{\text{man}} \Phi_{\text{man}}(x). \tag{9}$$

This formulation compares three levels of design control: unconstrained surrogate optimisation based on the first two terms of equation (9), physics-constrained optimisation with $\Phi_{\text{phys}}$, and topology-aware physics-constrained optimisation with both $\Phi_{\text{phys}}$ and $\Phi_{\text{man}}$.

No formal theorem is claimed from these equations. They are definitions of the design criteria used in the computational study. The mathematical role of each term is transparent: the prediction term measures target compliance, the physical term measures rule violation, and the manifold term measures distance from observed prototype regimes. The empirical question addressed later is whether adding the last two terms changes the quality and credibility of the generated candidates.

## 4 Dataset and design-space representation

The case study uses the public Mendeley Data benchmark of Pham (2023). The database includes 274 data points for GPC made with FA and GGBFS, with 19 input variables describing binder contents, aggregate contents, alkaline activators, water, superplasticizer, and curing conditions. Two targets are considered: 28-day compressive strength and carbon emission. The raw file provides separate sheets for strength and carbon emission. To avoid accidental misalignment, the two sheets are joined by the full 19-variable recipe and an occurrence counter for duplicated recipes.

The variables are listed in table 3. The notation $D0, \ldots, D18$ is retained because it is compact and consistent with the source file. Each symbol is introduced in the table before later use in feature definitions.

**Table 3:** Input variables.

| Symbol | Description | Symbol | Description |
|---|---|---|---|
| D0 | Fly ash, FA | D10 | Water in NaOH solution |
| D1 | Ground granulated blast-furnace slag, GGBFS | D11 | Dry NaOH |
| D2 | Coarse aggregate | D12 | Additional water |
| D3 | Fine aggregate | D13 | Superplasticizer |
| D4 | Sodium silicate, $Na_2SiO_3$ | D14 | Total water |
| D5 | Sodium hydroxide, NaOH | D15 | Initial curing time |
| D6 | Dry $Na_2O$ | D16 | Initial curing temperature |
| D7 | Dry $SiO_2$ | D17 | Initial curing rest time |
| D8 | Water in sodium silicate solution | D18 | Final curing temperature |
| D9 | NaOH concentration | | |

Several derived descriptors are used to make the design space more interpretable. Let $B$ denote binder content and $A$ denote alkaline activator content:

$$B = D0 + D1, \qquad A = D4 + D5. \tag{10}$$

The fly-ash fraction and slag fraction in the binder are

$$r_{\mathrm{FA}} = \frac{D0}{D0 + D1}, \qquad r_{\mathrm{GGBFS}} = \frac{D1}{D0 + D1}. \tag{11}$$

The activator-to-binder ratio, sodium-silicate-to-NaOH ratio, total-water-to-binder ratio, and superplasticizer-to-binder ratio are

$$r_{\mathrm{act/binder}} = \frac{D4 + D5}{D0 + D1}, \quad r_{\mathrm{sil/NaOH}} = \frac{D4}{D5}, \quad r_{\mathrm{water/binder}} = \frac{D14}{D0 + D1}, \quad r_{\mathrm{sp/binder}} = \frac{D13}{D0 + D1}. \tag{12}$$

Finally, two binary indicators are used to mark the presence of superplasticizer and additional water:

$$I_{\mathrm{sp}} = \mathbf{1}_{D13>0}, \qquad I_{\mathrm{aw}} = \mathbf{1}_{D12>0}. \tag{13}$$

These descriptors are not intended to replace the original variables. They provide interpretable coordinates for topology analysis, clustering, physical constraints, and inverse-design diagnostics.

A useful didactic interpretation is the following. The original 19 variables describe a recipe. The derived ratios describe the recipe in engineering terms. The topology analysis then asks whether these recipes fill the whole ambient design box or only a lower-dimensional feasible region. The inverse-design stage finally asks whether a proposed new recipe is close enough to that feasible region to be considered credible.

Before model fitting, all variables used in metric-based procedures are standardised on the training data only. The same transformation is then applied to validation or candidate points. This prevents scale-dominated distances and avoids leakage from validation folds. Derived ratios are computed after checking that their denominators are positive. Candidate recipes generated during inverse design are projected back to admissible ranges and discrete levels before final evaluation. These implementation details are important because small-data inverse design is sensitive to seemingly minor preprocessing choices.

# 5 Methodology

The methodology contains four connected components: topology analysis, surrogate prediction, INCRT-based rationalisation, and constrained inverse design. The implementation was carried out in Python using standard scientific-computing and machine-learning tools, including scikit-learn (Pedregosa et al., 2011) and XGBoost (Chen and Guestrin, 2016).

The components are intentionally separated. Topology analysis is unsupervised and describes the geometry of the recipes. Surrogate prediction is supervised and estimates strength or carbon emission. INCRT rationalisation converts the geometric information into prototypes and support scores. Constrained inverse design then uses these objects to screen candidates. This separation avoids mixing two different claims: a claim about prediction accuracy and a claim about design credibility.

**Topology and intrinsic dimensionality.** The purpose of the topology analysis is to determine whether the observed mixtures occupy the full 19-dimensional ambient input space or a lower-dimensional feasible subset. PCA is applied to the standardised representation. If $\lambda_1, \ldots, \lambda_d$ are the covariance eigenvalues, the cumulative explained variance is

$$\rho(q) = \frac{\sum_{j=1}^{q} \lambda_j}{\sum_{j=1}^{d} \lambda_j}. \tag{14}$$

The number of components needed to explain a fraction $\eta$ of variance is

$$d_\eta = \min\{q : \rho(q) \geq \eta\}. \tag{15}$$

The participation-ratio dimension and entropy effective rank are

$$d_{\mathrm{PR}} = \frac{\left(\sum_j \lambda_j\right)^2}{\sum_j \lambda_j^2}, \qquad d_{\mathrm{ER}} = \exp\left(-\sum_j q_j \log q_j\right), \qquad q_j = \frac{\lambda_j}{\sum_\ell \lambda_\ell}. \tag{16}$$

Local intrinsic dimensionality is estimated with the nearest-neighbour estimator of Levina and Bickel (2005). If $r_j(x_i)$ is the distance from point $x_i$ to its $j$-th nearest neighbour, the local estimate is

$$\widehat{d}_k(x_i) = \left[\frac{1}{k-1}\sum_{j=1}^{k-1} \log \frac{r_k(x_i)}{r_j(x_i)}\right]^{-1}. \tag{17}$$

Median values over all samples are reported for several neighbourhood sizes $k$. Clustering is then performed in the topology-aware representation, and the number of clusters is selected by silhouette score (Rousseeuw, 1987).

**Surrogate prediction.** Forward models are trained separately for compressive strength and carbon emission. The strength task is treated as nonlinear because strength depends on interacting chemical and curing variables. Tree ensembles are therefore included: Random Forests (Breiman, 2001), ExtraTrees (Geurts et al., 2006), Gradient Boosting (Friedman, 2001), and XGBoost (Chen and Guestrin, 2016). Support-vector regression is included as a kernel baseline (Cortes and Vapnik, 1995). Ridge and ElasticNet regression are included as regularised linear baselines, with ElasticNet following Zou and Hastie (2005). The carbon-emission task is treated separately because the target is composition-driven.

**INCRT-based rationalisation.** INCRT (Cirrincione, unpublished results) is used as a topology-aware rationalisation layer. In this paper, the INCRT layer is defined operationally by three objects: a set of prototypes $\mathcal{P}$, soft memberships to prototype regimes, and the manifold-support score $\Phi_{\mathrm{man}}$ in equation (8). This definition is sufficient for the design-support role needed here. It does not require INCRT to be the most accurate standalone regressor. Its purpose is to answer a different question: whether a candidate generated by inverse optimisation remains close to experimentally observed mixture regimes.

The connection to the original INCRT architecture is the following. INCRT is an attention-based architecture that grows incrementally one head at a time, driven by the directional structure of the data, until the residual signal that would trigger another head falls below a stopping criterion. At convergence, each active head is associated with a representative direction in the data, and the union of these directions induces a partition of the inputs into prototype-supported regimes. In the present paper, only this prototype partition and its associated regime centroids $\{p_k\}_{k=1}^K$ are retained. The full attention machinery, which is needed when INCRT is used as a sequence model, is not required for tabular mixture design and is therefore omitted. What survives is the self-organising prototype structure, which is precisely the object needed to define the manifold-support term $\Phi_{\mathrm{man}}$ in equation (8).

For a candidate $x$, a soft membership to prototype $p_k$ can be written as

$$u_k(x) = \frac{\exp\left[-d_{\mathrm{mix}}(x, p_k)/\tau\right]}{\sum_{\ell=1}^K \exp\left[-d_{\mathrm{mix}}(x, p_\ell)/\tau\right]}, \tag{18}$$

where $\tau > 0$ is a temperature parameter. These memberships provide local interpretability, while $\Phi_{\mathrm{man}}$ provides a scalar support score for optimisation.

**Physical constraints and candidate projection.** The physical penalty is implemented as a screening and penalisation device rather than as a complete chemical simulator. It includes positivity of material quantities, admissible bounds inherited from the observed data, consistency of derived ratios, and projection of discrete curing variables to admissible levels. This choice is deliberately conservative. The aim is not to prove that a generated candidate will necessarily succeed experimentally, but to reject candidates that are obviously inconsistent with the dataset and with basic mixture-design logic.

A candidate can fail in two different ways. It can be physically invalid, for example by violating a ratio or using an impossible level of a process variable. It can also be physically admissible but unsupported, meaning that it respects explicit rules while remaining far from all observed mixture regimes. The first failure is handled by $\Phi_{\text{phys}}$. The second failure is handled by $\Phi_{\text{man}}$. Both checks are needed because explicit rules cannot describe all hidden correlations present in a small experimental dataset.

**Validation protocol.** Repeated stratified cross-validation is used for ordinary predictive evaluation. Since the target is continuous, stratification is performed by quantile binning of strength. Nested cross-validation is used when hyperparameter tuning is required. Three metrics are reported: coefficient of determination $R^2$, root-mean-square error (RMSE), and mean absolute error (MAE). RMSE and MAE are defined as

$$\text{RMSE} = \sqrt{\frac{1}{n}\sum_{i=1}^{n}(y_i - \hat{y}_i)^2}, \qquad \text{MAE} = \frac{1}{n}\sum_{i=1}^{n}|y_i - \hat{y}_i|, \tag{19}$$

where $\hat{y}_i$ is the prediction for sample $i$.

Leave-one-cluster-out (LOCO) validation is used to test regime-shift generalisation. A model is trained on all but one topology-derived cluster and evaluated on the excluded cluster. This is stricter than random cross-validation because the test set corresponds to a coherent mixture regime absent from training.

The validation protocol is therefore deliberately redundant. Random repeated folds answer the question "how well is interpolation performed?" Nested selection reduces the risk of hyperparameter optimism. LOCO validation answers the question "what happens when an entire mixture regime is missing?" The three answers are not expected to coincide, and their discrepancy is itself informative for inverse design.

# 6 Results

## 6.1 Topology and design-space structure

The correlation analysis reveals that the 19 original descriptors are far from independent. Several variables belong to strongly coupled chemical or procedural groups. Sodium-silicate-related variables, NaOH-related variables, dry oxide contents, and water-related descriptors display strong dependencies. This confirms that the original representation contains substantial redundancy. The heatmap in figure 2 gives the global view, while the PCA spectrum quantifies the reduction.

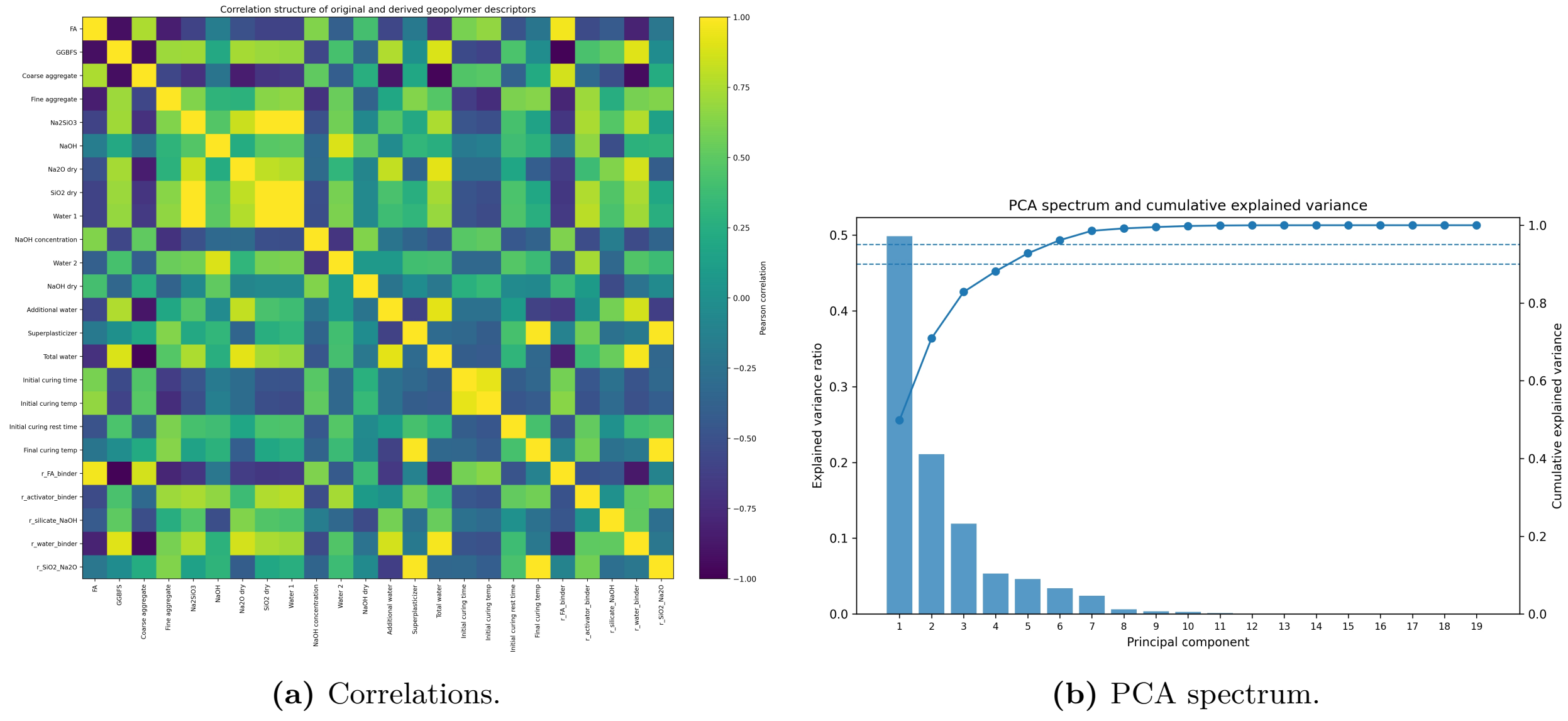

**(a)** Correlations. **(b)** PCA spectrum.

**Figure 2:** Redundancy diagnostics.

The first two principal components explain 71.2% of the input variance. Four components explain 88.8%, five components explain 93.4%, and six components explain 96.7%. Thus, the number of components required to explain 90% and 95% of the variance is $d_{90} = 5$ and $d_{95} = 6$, respectively. The participation-ratio dimension is $d_{\mathrm{PR}} = 3.17$, and the entropy effective rank is $d_{\mathrm{ER}} = 4.49$. Local intrinsic-dimensionality estimates provide an even lower local view: the median Levina-Bickel estimate decreases from about 3.01 for $k = 10$ to about 2.14 for $k = 30$. The complete summary is given in table 4 and visualised in figure 3.

**Table 4:** Effective dimensionality.

| Quantity | Value |
|---|---|
| PCs for 90% variance, $d_{90}$ | 5 |
| PCs for 95% variance, $d_{95}$ | 6 |
| Explained variance, 2 PCs | 71.2% |
| Explained variance, 4 PCs | 88.8% |
| Explained variance, 5 PCs | 93.4% |
| Explained variance, 6 PCs | 96.7% |
| Participation-ratio dimension | 3.17 |
| Entropy effective rank | 4.49 |
| Median local ID, $k = 10$ | 3.01 |
| Median local ID, $k = 20$ | 2.28 |
| Median local ID, $k = 30$ | 2.14 |

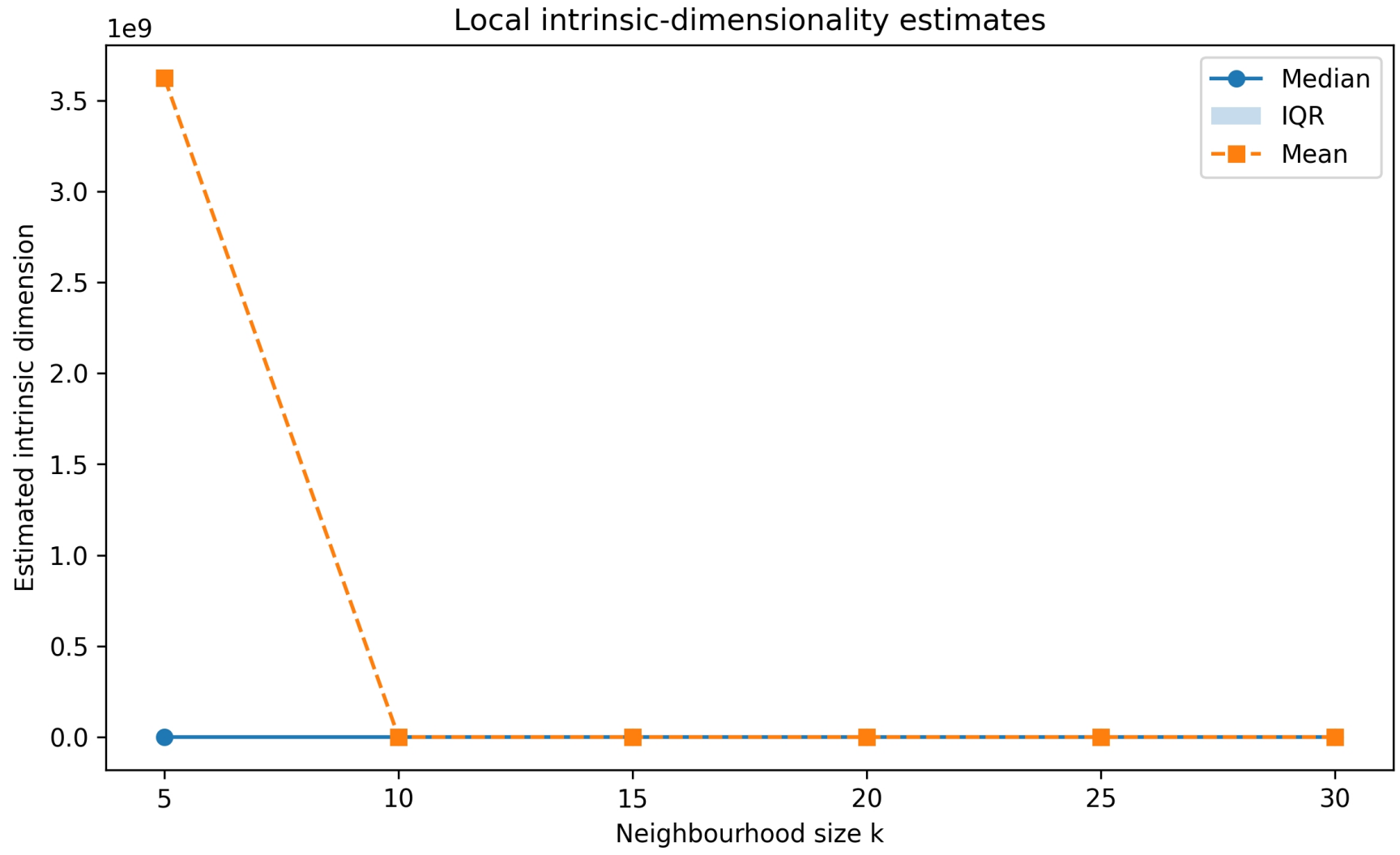

**Figure 3:** Local intrinsic dimension.

The topology-aware embedding and clustering analysis identify three main mixture regimes. The best tested cluster number is $K = 3$, with silhouette score 0.595. The clusters correspond to interpretable mixture families: a large mixed-binder regime, a fly-ash-rich regime, and a slag-rich/high-water regime. The cluster partition is shown in figure 4; its numerical summary is reported in table 5.

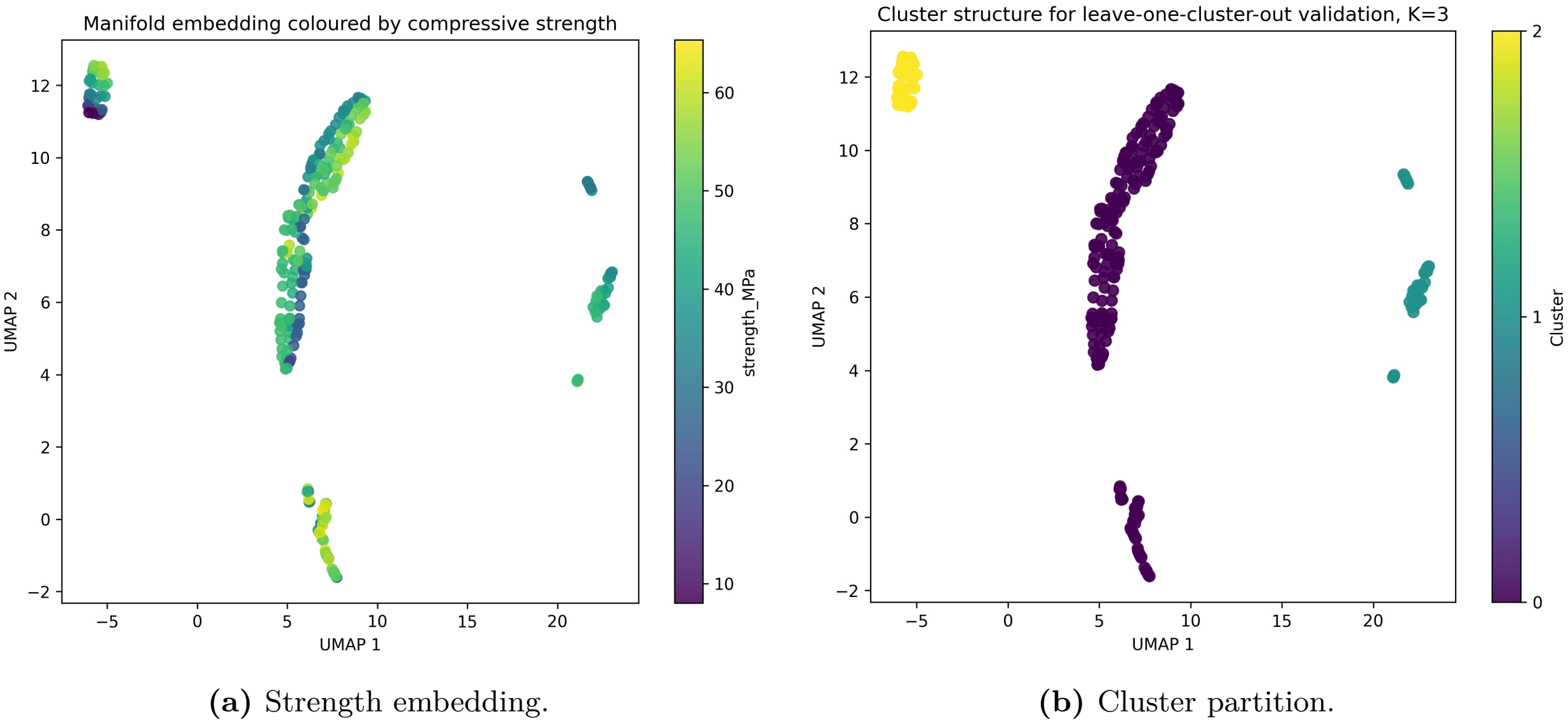

**(a)** Strength embedding. **(b)** Cluster partition.

**Figure 4:** Mixture regimes.

**Table 5:** Cluster summary.

| Cluster | Size | Mean strength | Mean $CO_2$ | FA/binder | Interpretation |
|---|---|---|---|---|---|
| 0 | 198 | 44.74 | 145.53 | 0.60 | Mixed-binder intermediate regime |
| 1 | 40 | 41.07 | 128.39 | 1.00 | Fly-ash-rich regime |
| 2 | 36 | 36.35 | 122.36 | 0.25 | Slag-rich/high-water regime |

These results support the first methodological premise of the paper. The 19-dimensional representation should not be interpreted as a generic design box. The observed mixtures occupy a structured lower-dimensional region. Consequently, inverse design in the full ambient domain requires a support-control mechanism.

The cluster interpretation should not be read as a definitive materials taxonomy. It is a data-driven partition of a small benchmark. Nevertheless, the regimes are chemically meaningful enough to be useful: one cluster is dominated by mixed binders, one by fly ash, and one by a slag-rich/high-water configuration. This level of interpretability is sufficient for the purpose of LOCO testing and for defining prototype-supported neighbourhoods.

## 6.2 Forward prediction

The repeated stratified cross-validation results for 28-day compressive strength are reported in table 6. The best predictive models are tree ensembles and boosting methods. ExtraTrees gives the best average result, followed closely by XGBoost, Gradient Boosting, and Random Forest. Linear models are substantially weaker, which confirms the nonlinear nature of the strength target.

**Table 6:** Strength prediction.

| Model | $R^2$ | RMSE [MPa] | MAE [MPa] |
|---|---|---|---|
| ExtraTrees | **0.951 ± 0.018** | **2.48 ± 0.38** | **1.40 ± 0.26** |
| XGBoost | 0.947 ± 0.016 | 2.59 ± 0.35 | 1.88 ± 0.21 |
| GradientBoosting | 0.947 ± 0.016 | 2.60 ± 0.35 | 1.88 ± 0.22 |
| RandomForest | 0.942 ± 0.021 | 2.71 ± 0.41 | 1.88 ± 0.28 |
| SVR-RBF | 0.886 ± 0.028 | 3.81 ± 0.37 | 2.76 ± 0.32 |
| kNN-kernel | 0.836 ± 0.054 | 4.55 ± 0.67 | 2.97 ± 0.38 |
| Ridge | 0.790 ± 0.040 | 5.19 ± 0.39 | 4.40 ± 0.29 |
| INCRT-ensemble | 0.660 ± 0.051 | 6.63 ± 0.45 | 5.34 ± 0.48 |
| INCRT-local-ridge | 0.649 ± 0.061 | 6.72 ± 0.51 | 5.40 ± 0.55 |

The prototype-only INCRT regressors do not outperform tree-based models in forward prediction. This observation is important. It indicates that INCRT should not be presented as the best standalone regressor for this dataset. Its more appropriate role is topology-aware rationalisation and manifold support. The strongest tabular surrogate is used for prediction; INCRT is used to decide whether a candidate is supported by the learned design manifold.

This result also prevents a common overclaim in model-driven materials papers. The framework does not require every component to be best at every task. The predictive surrogate is selected for numerical accuracy, while the topology layer is selected for support assessment. A less accurate but interpretable prototype structure can still be valuable if it prevents the optimiser from proposing recipes that are far from all experimentally observed regimes.

The carbon-emission target behaves differently from compressive strength. The results in table 7 show that regularised linear models recover carbon emission almost exactly. ElasticNet gives $R^2 = 0.9999 \pm 0.0001$, with very small RMSE and MAE. This indicates that carbon emission is essentially composition-driven in this benchmark, whereas strength remains the nonlinear prediction task.

**Table 7:** Carbon-emission prediction.

| Model | $R^2$ | RMSE | MAE |
|---|---|---|---|
| ElasticNet | **0.9999 ± 0.0001** | **0.194 ± 0.049** | **0.128 ± 0.022** |
| Ridge | 0.9993 ± 0.0004 | 0.485 ± 0.109 | 0.301 ± 0.050 |
| SVR-RBF | 0.993 ± 0.008 | 1.44 ± 0.65 | 0.59 ± 0.21 |
| ExtraTrees | 0.984 ± 0.010 | 2.41 ± 0.69 | 1.40 ± 0.31 |
| XGBoost | 0.980 ± 0.010 | 2.68 ± 0.60 | 1.71 ± 0.32 |

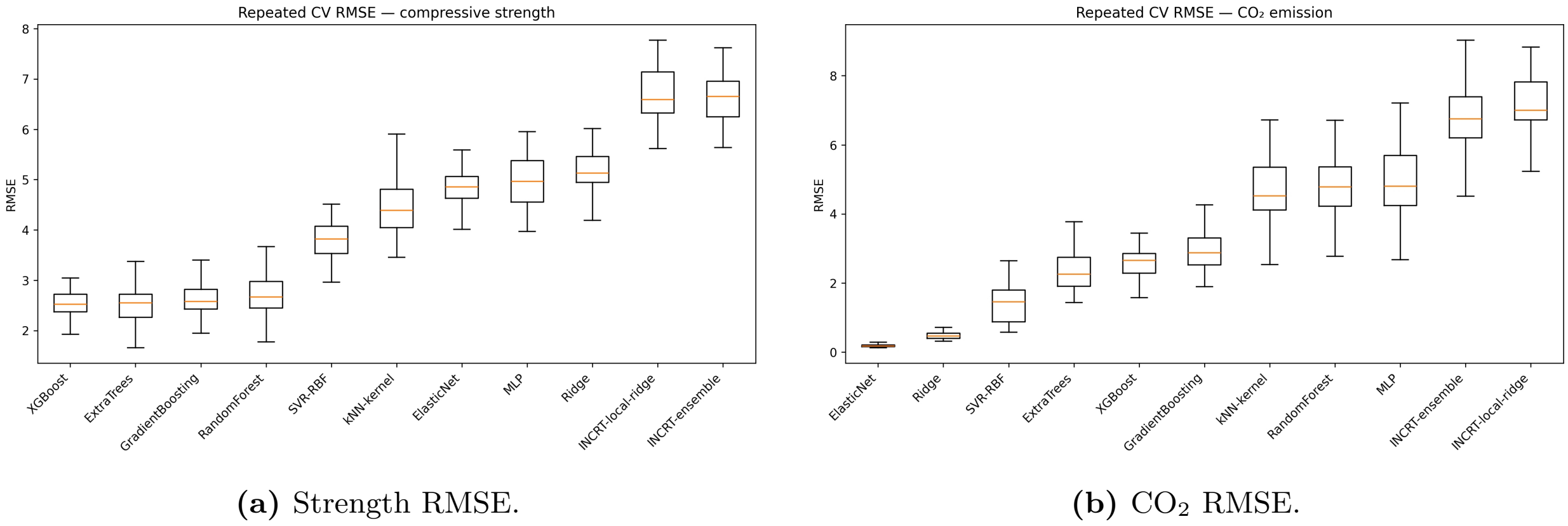

**(a)** Strength RMSE. **(b)** $CO_2$ RMSE.

**Figure 5:** Forward-prediction errors.

The LOCO results reveal a much harder regime-shift problem. Under repeated stratified cross-validation, ExtraTrees obtains an RMSE of approximately 2.48 MPa. Under LOCO validation, its strength RMSE increases to approximately 14.20 MPa. The degradation is shown in figure 6. This result confirms that random cross-validation evaluates interpolation, whereas cluster-level validation reveals the difficulty of generalising to an unseen mixture regime.

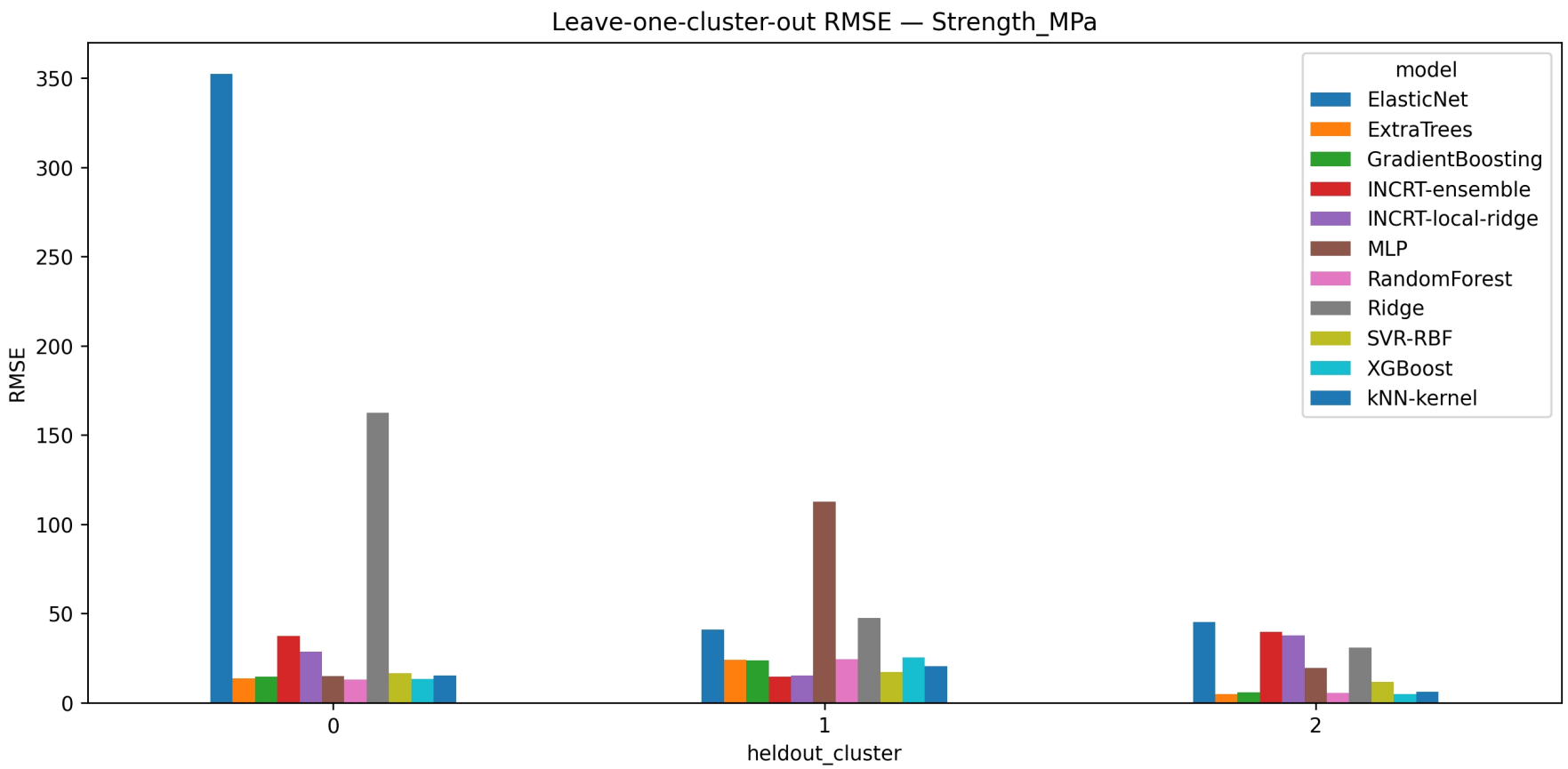

**Figure 6:** LOCO strength errors.

The practical consequence is direct: if surrogate models degrade under cluster-level regime shift, inverse candidates located far from the observed manifold cannot be trusted merely because the surrogate predicts attractive performance. The inverse-design stage must therefore include a manifold-support term.

A useful way to read the LOCO experiment is as a stress test. It does not claim that future candidates will necessarily be identical to a missing cluster. Instead, it shows that leaving the familiar interpolation regime can change the error scale by several multiples. This makes a purely surrogate-optimal candidate less convincing unless its position in the design space is also examined.

### 6.3 Physics-constrained inverse design

The final inverse-design system uses ExtraTrees for strength prediction, ElasticNet for carbon-emission prediction, and the INCRT/topology layer for manifold support. The topology-aware representation uses a six-dimensional PCA/prototype manifold. The first six principal components explain 95.77% of the variance in the augmented design representation. The empirical 95th and

99th percentiles of training-sample distances to the prototype set are

$$Q_{95} = 6.51, \qquad Q_{99} = 11.38. \tag{20}$$

Candidates with $\Phi_{\text{man}}(x) < 6.51$ are interpreted as well supported. Candidates with $6.51 < \Phi_{\text{man}}(x) < 11.38$ are interpreted as borderline but inside the empirical 99% support region. Candidates with $\Phi_{\text{man}}(x) > 11.38$ are interpreted as off-manifold.

The inverse-design experiments were performed for target strengths of 30, 40, 50, and 60 MPa. Strategy A denotes unconstrained surrogate optimisation. Strategy B denotes physics-constrained optimisation. Strategy C denotes topology-aware physics-constrained optimisation. The comparison is reported in table 8 and visualised in figure 7.

**Table 8:** Inverse-design comparison.

| Target | Strategy | Pred. strength | Error | Pred. $CO_2$ | Phys. viol. | Manifold | Interpretation |
|---|---|---|---|---|---|---|---|
| 30 | A | 30.00 | 0.00 | 130.55 | 0 | 1.65 | exact target, not low-carbon |
| 30 | B | 30.00 | 0.00 | 111.21 | 0 | 41.29 | physically valid, off-manifold |
| 30 | C | 33.90 | 3.90 | **85.99** | 0 | 3.74 | low-carbon, supported |
| 40 | A | 39.99 | 0.01 | 94.96 | 1 | 38.84 | violation and off-manifold |
| 40 | B | 40.00 | 0.00 | 135.38 | 0 | 4.68 | valid, higher carbon |
| 40 | C | 42.04 | 2.04 | **78.02** | 0 | 11.04 | low-carbon, borderline |
| 50 | A | 50.00 | 0.00 | 135.85 | 1 | 1.13 | target match, violation |
| 50 | B | 50.00 | 0.00 | 125.67 | 0 | 5.12 | valid, supported |
| 50 | C | 50.03 | **0.03** | **113.16** | 0 | 2.71 | best balance |
| 60 | A | 60.00 | 0.00 | 135.14 | 1 | 8.90 | target match, violation |
| 60 | B | 60.00 | 0.00 | 169.85 | 0 | 4.04 | valid, high carbon |
| 60 | C | 58.33 | 1.67 | **115.26** | 0 | 2.50 | low-carbon trade-off |

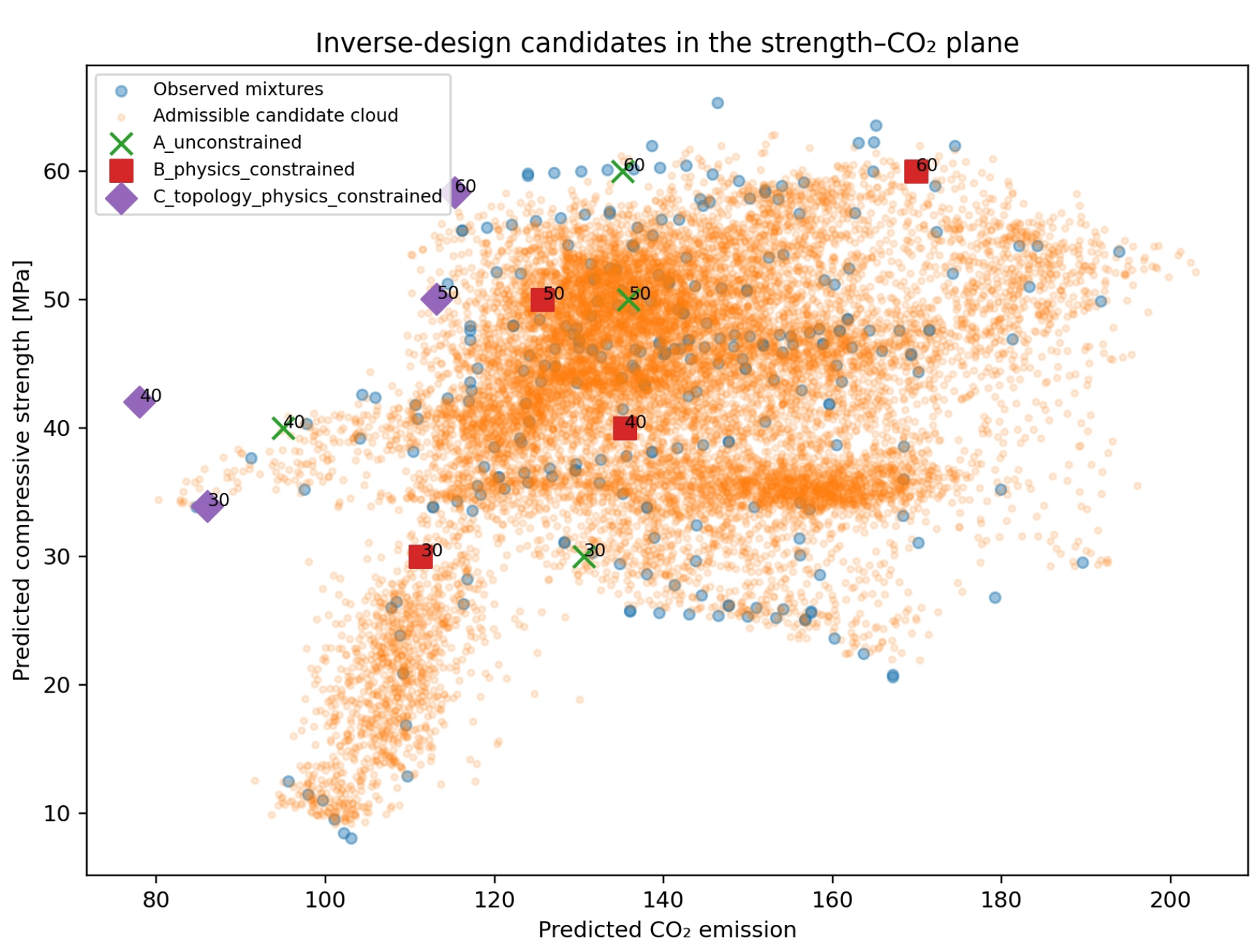

**Figure 7:** Inverse-design candidates.

The table shows that exact target matching is not sufficient to define a useful engineering design. Strategy A often reaches the target almost exactly, but it can produce candidates with physical violations or off-manifold behaviour. Strategy B removes physical violations, but it can still produce candidates with poor manifold support or unnecessarily high carbon emission. Strategy C sacrifices exact target matching in some cases, but provides candidates that are physically admissible, lower in carbon emission, and supported by the learned design manifold.

The 50 MPa and 60 MPa cases are particularly informative. For the 50 MPa target, the topology-aware candidate reaches 50.03 MPa with a target error of only 0.03 MPa, predicted carbon emission of 113.16, zero physical violations, and low manifold score. For the 60 MPa target, the physics-constrained candidate reaches the target exactly but has predicted carbon emission of 169.85. The topology-aware candidate gives 58.33 MPa and 115.26 predicted carbon emission. A small strength tolerance therefore produces a large carbon reduction while preserving physical feasibility and manifold support.

For the 30 MPa and 40 MPa targets, the topology-aware strategy should be interpreted as a carbon-aware design strategy under strength tolerance, not as an exact target interpolator. At 30 MPa, Strategy C accepts a moderate strength overshoot in exchange for much lower carbon emission and good manifold support. At 40 MPa, Strategy C gives a low-carbon candidate whose manifold score remains below the empirical 99% threshold but close to the boundary. This distinction is important because it avoids overclaiming exact target optimality.

The inverse-design table should therefore be read column by column rather than by target error alone. A candidate with zero error can still be unattractive if it violates physical constraints or has high carbon emission. A physically valid candidate can still be risky if it is far from the data manifold. A topology-aware candidate can be preferable even with a small target deviation, provided that the deviation is acceptable for screening and the gains in carbon emission and support are substantial.

# 7 Discussion

The results support the central premise of the study: in small-data engineering inverse design, prediction accuracy is necessary but not sufficient. The GPC benchmark can be predicted accurately under repeated random cross-validation, especially for compressive strength when tree-based ensemble models are used. However, the broader objective is not only forward prediction. The central issue is whether a data-driven model can support reliable inverse design from a small, mixed, physically constrained tabular dataset.

The low-dimensionality results are important in this respect. A dataset with 274 samples would be very sparse in a truly unconstrained 19-dimensional space. The observed PCA, effective-rank, and local intrinsic-dimensionality results indicate that the effective structure is much smaller. This makes prediction feasible, but it also changes the inverse-design problem. If feasible mixtures occupy a thin subset of the ambient space, optimisation over the full variable box is risky. A candidate may satisfy all individual bounds but still be unrealistic because it violates the learned low-dimensional structure of the data.

The different behaviour of strength and carbon-emission prediction is also informative. Carbon emission is almost perfectly recovered by regularised linear models, indicating that it is largely determined by mixture composition. Strength is more difficult because it depends on nonlinear interactions among precursor composition, activator chemistry, water content, curing temperature, curing time, and other mixture variables. This justifies the final hybrid architecture: a nonlinear tree ensemble for strength, a regularised linear model for carbon emission, and INCRT prototypes for manifold support.

The LOCO result explains why random cross-validation should not be the only validation protocol. Repeated stratified cross-validation mostly tests interpolation among related mixtures. LOCO validation tests whether a coherent mixture regime can be predicted when it is absent from training. The increase of ExtraTrees RMSE from approximately 2.48 MPa to approximately 14.20 MPa indicates that regime-shift generalisation remains difficult. Inverse design is precisely the situation where such regime shifts can occur, because an optimiser can move toward the edge of known regimes.

The role of INCRT is clarified by the experiments. Prototype-only INCRT regressors are not the most accurate forward predictors on this dataset. This does not invalidate the INCRT component. Instead, it shows that INCRT should be used for what it does best in this setting: prototype rationalisation, local regime description, manifold support, and out-of-distribution control. This avoids an unnecessary competition between INCRT and highly optimised tabular predictors.

This point is important for editorial clarity. The claim is not that an INCRT regressor dominates established tabular learners. The claim is that inverse design needs a support mechanism in addition to a predictor, and that an INCRT-style prototype layer provides such a mechanism in a transparent form. The final framework is therefore more defensible:

| accurate surrogate prediction + INCRT manifold rationalisation + physics-constrained inverse design |
|---|

Several limitations must be acknowledged. First, the dataset contains only 274 mixtures. Although the low-dimensional structure makes learning feasible, candidate mixtures generated by inverse design should be regarded as hypotheses for experimental validation, not as final certified recipes. Second, the carbon-emission target appears almost deterministic from input composition; this is useful for optimisation, but it is not comparable in difficulty to strength prediction. Third, the physical constraints used here are simplified. They include bounds, admissible ratios, and projection mechanisms, but they do not represent a complete chemical model of geopolymerisation. More detailed constraints could include workability, setting time, molar ratios, raw-material availability, curing energy, and durability indicators. Fourth, the manifold-support score is data-driven. It reduces extrapolation risk, but it does not prove experimental feasibility.

A further limitation concerns uncertainty. The present study uses repeated validation, LOCO testing, and manifold support to expose risk, but it does not provide a calibrated predictive interval for each generated recipe. In a deployment setting, a candidate should ideally be accompanied by both a predicted mean and a reliable uncertainty measure. The manifold score can flag unsupported regions, but it is not equivalent to a probabilistic confidence statement. This distinction should be preserved when the framework is used for experimental planning.

Future work should combine this framework with active learning. The manifold score and predictive uncertainty could be used to select new experiments that maximally improve the design space, especially in high-strength low-carbon regions. The inverse-design stage could also be extended to additional objectives, including cost, workability, durability, curing energy, and thermal conductivity. Uncertainty quantification should be strengthened through ensemble uncertainty, conformal prediction, or Bayesian surrogates. Finally, experimental validation remains necessary. The most promising topology-aware candidates, especially those around 50 and 60 MPa, should be tested in the laboratory.

# 8 Conclusion

A topology-aware inverse-design framework has been developed for small, mixed, physically constrained geopolymer mixture data. The study started from a practical difficulty: good forward prediction does not automatically imply reliable inverse design. In a small dataset, a numerical optimiser can easily move outside experimentally supported regions, even when surrogate predictions remain attractive.

The empirical results show that the 19-variable GPC representation is highly redundant. Five principal components explain more than 93% of the variance, and local intrinsic-dimensionality estimates indicate local mixture regimes of approximately two to three dimensions. This supports the interpretation of the dataset as a low-dimensional feasible manifold embedded in a higher-dimensional tabular space.

The prediction results separate the two targets clearly. Compressive strength is nonlinear and is best handled by tree-based ensemble surrogates. Carbon emission is almost composition-determined and is accurately recovered by regularised linear models. This observation motivates a hybrid modelling strategy rather than a single universal predictor.

The inverse-design results show why topology matters. Unconstrained surrogate optimisation can hit target strength but may violate physical constraints or leave the data-supported manifold. Physics-only optimisation removes explicit violations but does not always prevent off-manifold or high-carbon solutions. The topology-aware strategy provides a more cautious design filter: candidate mixtures are selected by considering target error, carbon emission, physical admissibility, and manifold support simultaneously.

The clearest benefits are observed in the 50 MPa and 60 MPa cases. At 50 MPa, the topology-aware candidate nearly matches the target while reducing predicted carbon emission relative to the physics-only candidate. At 60 MPa, a small strength tolerance produces a large predicted carbon reduction while maintaining good manifold support. These results indicate that exact target matching is not the only relevant criterion in sustainable mixture design.

The proposed framework is not a substitute for experimental validation. It should be interpreted as a decision-support methodology for screening and ranking credible candidate mixtures before laboratory testing. Its main value lies in reducing the risk of surrogate artefacts by combining accurate prediction, physical admissibility, and data-supported topology. This principle is not specific to geopolymers and can be transferred to other small-data engineering inverse-design problems.

The practical conclusion is simple. A recommended mixture should be accompanied by four pieces of information: predicted performance, predicted carbon emission, physical feasibility status, and manifold-support status. Without the last two, inverse design remains vulnerable to attractive but unsupported surrogate optima. With them, the output becomes a ranked set of experimentally testable hypotheses rather than a list of unqualified numerical optima.

## Acronyms

Acronyms used in the paper.

| Acronym | Meaning |
|---|---|
| AEI | Advanced Engineering Informatics |
| $CO_2$ | Carbon dioxide |
| FA | Fly ash |
| GGBFS | Ground granulated blast-furnace slag |
| GPC | Geopolymer concrete |
| INCRT | Incremental Transformer |
| LOCO | Leave-one-cluster-out |
| MAE | Mean absolute error |
| OOD | Out-of-distribution |
| OPC | Ordinary Portland cement |
| PCA | Principal component analysis |
| RMSE | Root-mean-square error |
| SVR | Support-vector regression |

## References

Leo Breiman. Random forests. *Machine Learning*, 45(1):5–32, 2001. doi: 10.1023/A:1010933404324.

Tianqi Chen and Carlos Guestrin. XGBoost: A scalable tree boosting system. In *Proceedings of the*

*22nd ACM SIGKDD International Conference on Knowledge Discovery and Data Mining*, pages 785–794. ACM, 2016. doi: 10.1145/2939672.2939785.

Giansalvo Cirrincione. INCRT: An incremental transformer that determines its own architecture. arXiv preprint arXiv:2604.10703, unpublished results. URL https://arxiv.org/abs/2604.10703.

Corinna Cortes and Vladimir Vapnik. Support-vector networks. *Machine Learning*, 20:273–297, 1995. doi: 10.1007/BF00994018.

Joseph Davidovits. *Geopolymer Chemistry and Applications*. Geopolymer Institute, Saint-Quentin, France, 4 edition, 2015. ISBN 978-2-9514820-9-8. 623 pages.

Jerome H. Friedman. Greedy function approximation: A gradient boosting machine. *The Annals of Statistics*, 29(5):1189–1232, 2001. doi: 10.1214/aos/1013203451.

Pierre Geurts, Damien Ernst, and Louis Wehenkel. Extremely randomized trees. *Machine Learning*, 63:3–42, 2006. doi: 10.1007/s10994-006-6226-1.

Ian T. Jolliffe. *Principal Component Analysis*. Springer, New York, 2 edition, 2002. ISBN 978-0-387-95442-4. 488 pages.

Elizaveta Levina and Peter J. Bickel. Maximum likelihood estimation of intrinsic dimension. In *Advances in Neural Information Processing Systems 17*, pages 777–784. MIT Press, 2005.

Mohammad Mohsin, Tanima Ghosh, and Nusrat Hoque. Prediction and optimization of strength and CO2 emission for geopolymer concrete mix design using machine learning. *Results in Materials*, 28:100791, 2025. doi: 10.1016/j.rinma.2025.100791.

Ho Anh Thu Nguyen, Duy Hoang Pham, Yonghan Ahn, Bee Lan Oo, and Benson Teck Heng Lim. Machine learning and sustainable geopolymer materials: A systematic review. *Materials Today Sustainability*, 30:101095, 2025. doi: 10.1016/j.mtsust.2025.101095.

Fabian Pedregosa, Gael Varoquaux, Alexandre Gramfort, Vincent Michel, Bertrand Thirion, Olivier Grisel, Mathieu Blondel, Peter Prettenhofer, Ron Weiss, Vincent Dubourg, Jake Vanderplas, Alexandre Passos, David Cournapeau, Matthieu Brucher, Matthieu Perrot, and Édouard Duchesnay. Scikit-learn: Machine learning in python. *Journal of Machine Learning Research*, 12: 2825–2830, 2011.

Van Ngoc Pham. Compressive strength of geopolymer concrete. Mendeley Data, Version 1, 2023. Dataset, doi: 10.17632/36m95n7294.1.

Van-Ngoc Pham and Van-Quang Nguyen. Effects of alkaline activators and other factors on the properties of geopolymer concrete using industrial wastes based on gep-based models. *European Journal of Environmental and Civil Engineering*, 28(16):3770–3792, 2024. doi: 10.1080/19648189.2024.2357677.

John L. Provis and Jannie S. J. van Deventer, editors. *Geopolymers: Structures, Processing, Properties and Industrial Applications*. Woodhead Publishing, Cambridge, UK, 2009. ISBN 978-1-84569-449-4. 464 pages.

M. Srinivasula Reddy, Dinakar Pasla, and B. Hanumantha Rao. A review of the influence of source material's oxide composition on the compressive strength of geopolymer concrete. *Microporous and Mesoporous Materials*, 234:12–23, 2016. doi: 10.1016/j.micromeso.2016.07.005.

Peter J. Rousseeuw. Silhouettes: A graphical aid to the interpretation and validation of cluster analysis. *Journal of Computational and Applied Mathematics*, 20:53–65, 1987. doi: 10.1016/0377-0427(87)90125-7.

Olivier Roy and Martin Vetterli. The effective rank: A measure of effective dimensionality. In *Proceedings of the 15th European Signal Processing Conference*, pages 606–610, Poznan, Poland, 2007.

Christoph Völker, Benjamí Moreno Torres, Tehseen Rug, Rafia Firdous, Ghezal Ahmad Jan Zia, Stefan Lüders, Horacio Lisdero Scaffino, Michael Höpler, Felix Böhmer, Matthias Pfaff, Dietmar Stephan, and Sabine Kruschwitz. Data driven design of alkali-activated concrete using sequential learning. *Journal of Cleaner Production*, 418:138221, 2023. doi: 10.1016/j.jclepro.2023.138221.

Hui Zou and Trevor Hastie. Regularization and variable selection via the elastic net. *Journal of the Royal Statistical Society: Series B*, 67(2):301–320, 2005. doi: 10.1111/j.1467-9868.2005.00503.x.